\pdfoutput=1

\documentclass[11pt]{article}

\usepackage[]{acl}

\usepackage{times}
\usepackage{latexsym}
\usepackage{graphicx}
\usepackage{enumitem}
\usepackage{booktabs}
\usepackage{multirow}
\usepackage[T1]{fontenc}

\usepackage[utf8]{inputenc}
\usepackage{amsmath,amsfonts,amssymb,bm}
\renewcommand{\vec}[1]{\boldsymbol{#1}}
\usepackage{xcolor}
\usepackage{microtype}
\usepackage{pifont}
\usepackage{tcolorbox}
\usepackage{subfigure}

\newcommand{\dataset}{\textsc{DFlow}}

\newcommand{\datasets}{\textsc{DFlow }}

\usepackage{caption}

\usepackage{colortbl}

\usepackage{hyperref}
\usepackage{url}
\usepackage{dirtytalk}

\usepackage{adjustbox}
\usepackage{array}
\usepackage{booktabs}

\usepackage{enumitem}
\setlist[itemize]{leftmargin=*}
\setlist[itemize]{noitemsep}
\usepackage{etoolbox}
\makeatletter
\preto{\@tabular}{\parskip=0pt}
\makeatother

\usepackage{placeins}

\usepackage{titlesec}

\title{\underline{\dataset}: \underline{D}iverse Dialogue \underline{Flow} Simulation with Large Language Models}

\author{
    Wanyu Du\Thanks{~Work performed during an internship at AWS AI Labs.}\quad Song Feng\quad James Gung\quad \\\bf{Lijia Sun \quad Yi Zhang\quad Saab Mansour\quad Yanjun Qi} \\
    Amazon Web Services\\
    \texttt{\{dwanyu, sofeng, gungj\}@amazon.com} \\
    \texttt{\{sunlijia, yizhngn, saabm, yanjunqi\}@amazon.com}
}

\begin{document}
\maketitle

\begin{abstract}

Developing language model-based dialogue agents requires effective data to train models that can follow specific task logic. 
However, most existing data simulation methods focus on increasing diversity in language, topics, or dialogue acts at the utterance level, largely neglecting a critical aspect of task logic diversity at the dialogue level. 
This paper proposes a novel data simulation method designed to enhance the diversity of synthetic dialogues by focusing on task execution logic.
Our method uses LLMs to generate decision tree-structured task plans, which enables the derivation of diverse dialogue trajectories for a given task. Each trajectory, referred to as a "dialog flow", guides the generation of a multi-turn dialogue that follows a unique  trajectory. 
We apply this method to generate a task-oriented dialogue dataset comprising 3,886 dialogue flows across 15 different domains. We validate the effectiveness of this dataset using the next action prediction task, where models fine-tuned on our dataset outperform strong baselines, including GPT-4. 
Upon acceptance of this paper, we plan to release the code and data publicly.

\end{abstract}

\begin{figure*}[t]
    \centering
    \includegraphics[width=0.98\textwidth]{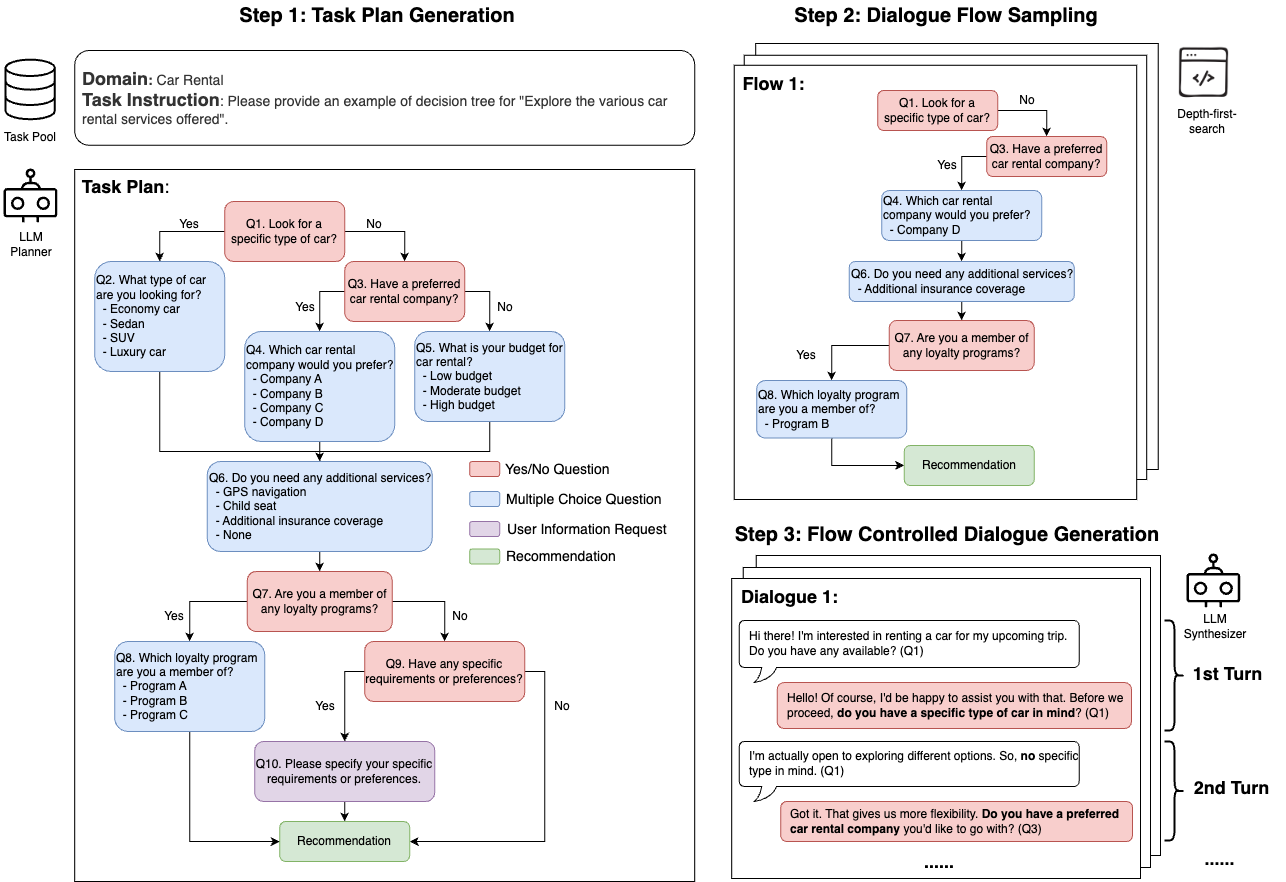}
    \caption{The proposed task-oriented dialogue data simulation framework. Given a task instruction, we use an LLM planner to generate a task plan $\vec{x}_{\text{plan}} \sim p(\vec{x}_{\text{plan}} \mid \vec{x}_{\text{goal}})$ in the format of a decision tree. 
Then depth-first-search is applied to parse all valid paths from the tree as dialogue flows. 
The dialogue flow is further used to control the LLM synthesizer to generate a multi-turn dialogue $\vec{x}_{\text{dialog}} \sim p(\vec{x}_{\text{dialog}} \mid \vec{x}_{\text{flow}})$. }
    \label{fig:overview}
    \vspace{-2mm}
\end{figure*}
\section{Introduction}

A task-oriented LLM agent typically needs to manage user requests according to a designated "plan guide" which includes predefined task logics and policy constraints. This guideline could indicate different task trajectories or flows that lead to task completions. For example, when exploring a car rental service, the agent will first ask if the user has a preferred rental company, if yes, then the agent will further let the user to specify the preferred company from a constrained list. Previous research \citep{mosig2020star,chen-etal-2021-action,raghu-etal-2021-end} has introduced dialogue flows to enable dialogue agents to follow task logic and constraints when addressing user requests. However, manually curating such dialogue flows is challenging due to the intricate task logic and alignment with policy constraints across various domains. Consequently, existing task-oriented dialogue datasets often suffer from sparse flow annotations, and lack efficient and generalizable mechanisms for automated dialogue flow generation \citep{budzianowski-etal-2018-multiwoz,byrne-etal-2019-taskmaster,rastogi2020towards,shalyminov2020fast,hattami2023workflow}. 

In this work, we aim at designing an automatic data simulation method to generate task-oriented dialogues with structured dialogue flows. Previous studies have utilized LLMs to generate task-oriented dialogues, focusing primarily on enhancing linguistic diversity \citep{li-etal-2022-controllable}, exploring varied topics \citep{xu-etal-2023-baize, ding-etal-2023-enhancing}, or proposing different combinations of dialogue acts at the utterance level \citep{chen-etal-2023-mixture}. However, these efforts neglect the crucial need for maintaining task logic consistency at the dialogue level. Our work proposes to use LLMs to generate diverse dialogues that consistently follow the task logic and constraints defined by dialogue flows. 

To achieve this goal, we propose a dialogue simulation framework that leverages the planning capabilities of LLMs \citep{yao2023tree, wang-etal-2023-plan} to automatically construct diverse dialogue flows. Starting with a task instruction, our framework employs an LLM planner to generate a decision tree-structured task plan that outlines diverse trajectories. Then depth-first-search is applied to parse all valid trajectories within this plan. Each trajectory represents a dialogue flow, ensuring a coherent task logic to effectively solve the given task. Subsequently, our framework uses the simulated dialogue flows to control an LLM synthesizer to generate multi-turn dialogues that follow task logics at the dialogue level. As shown in \autoref{fig:overview}, the LLM planner can simulate diverse dialogue flows, and the LLM synthesizer can generate coherent dialogues based on different dialogue flows.

To evaluate the effectiveness of the proposed framework, we construct a synthetic dataset \datasets and conduct extensive empirical experiments. Intrinsic evaluation on \datasets shows that this synthetic dataset obtains high quality task plans, coherent flows and faithful dialogues across 130 tasks in 15 domains.
Extrinsic evaluation demonstrates that 7B language models fine-tuned on \datasets dataset can outperform strong LLMs such as GPT-4 \cite{gpt_4} in the next action prediction task \citep{hattami2023workflow}.

In summary, this work introduces a novel data simulation method to synthesize diverse multi-turn dialogues that follow task logic and constraints. The simulated dataset enhances the dialogue understanding capabilities of small language models, enabling them to outperform more advanced LLMs such as GPT4 in the next action prediction task.

\section{Related Work}
\label{sec:related}


\paragraph{Dialogue Simulation with LLMs.}
Many prior works leverage LLMs to simulate dialogue datasets. 
\citet{li-etal-2022-controllable} propose to prompt GPT-3 with retrieved dialogue examples from existing datasets to generate new dialogues in a controllable way. 
However, the diversity of the generated dialogues is constrained by the retrieved dialogue examples. 
\citet{chen-etal-2023-mixture} design a soft-prompt tuning method to create a diverse set of prompts to synthesize diverse utterances using LLMs. 
But they only promote diversity at the utterance-level, ignoring the task logic at the dialogue-level.
Other works \citep{wang-etal-2023-self-instruct,ding-etal-2023-enhancing,chan2024scaling} propose to use the LLM-generated knowledge texts to synthesize diverse dialogues.
However, the knowledge text does not decompose the complex task into a step-by-step plan, making the dialogue generation process less controllable.

\paragraph{Task Planning with LLMs.}
Recent works enhance the planning ability of LLMs to solve complex tasks.
\citet{yao2023tree} proposes a tree-of-thought prompt to frame the problem solving as a search over a tree, and design search algorithms to obtain the best solution path to the problem.
\citet{wang-etal-2023-plan} designs a plan-and-solve prompting method, which generates a plan to divide the entire task into smaller subtasks.
\citet{liu2024logic} introduce a logic-of-thought prompt that employs propositional logic to generate expanded logical information from input context, further enhancing the logical reasoning capability of LLMs.
In this work, we leverage the task planning ability of LLMs to generate tree-structured task plans.
In contrast with enhancing the planning ability to find the optimal task trajectory, this work focuses on exploring diverse task trajectories that cover diverse task scenarios.

\begin{figure*}[h]
    \centering
    \begin{tcolorbox}[colback=gray!10,colframe=black,fontupper=\small,title=Prompt for Dialogue Generation with Dialogue Flow]
\textbf{Input}: Given a task flow:\\
    ```\\
    1. Are you looking for a specific type of car? - No.\\
    3. Do you have a preferred car rental company? - Yes.\\
    4. Which car rental company would you prefer? - Company D.\\
    6. Do you need any additional services? - Additional insurance coverage.\\
    7. Are you a member of any loyalty programs? - Yes.\\
    8. Which loyalty program are you a member of? - Program B.\\
    Recommendation\\
    ```\\
    Please create a natural conversation between a user and an agent using the given task flow. \\
    Include the question number at the end of each utterance. \\  
     \textbf{Output}:  \\
     \textit{User}: Hi there! I'm interested in renting a car for my upcoming trip. Do you have any available? (Question 1)\\
    \textit{Agent}: Hello! Of course, I'd be happy to assist you with that. Before we proceed, do you have a specific type of car in mind, or are you open to exploring different options? (Question 1)\\
    \textit{User}: I'm actually open to exploring different options. So, no specific type in mind. (Question 1)\\
    \textit{Agent}: Got it. That gives us more flexibility. Do you have a preferred car rental company you'd like to go with? (Question 3)\\
    \textit{User}: Yes, I've had good experiences with Company D in the past, so I'd prefer to rent from them. (Question 4)\\    
    \textit{Agent}: Great choice! Company D is reliable. Apart from the rental itself, do you need any additional services, like insurance coverage or anything else? (Question 6)\\
    \textit{User}: Yes, I would definitely like to have additional insurance coverage for peace of mind during my trip. (Question 6)\\
    \textit{Agent}: Safety first! I see you're a savvy traveler. By the way, are you a member of any loyalty programs? (Question 7)\\
    \textit{User}: Absolutely! Yes, I'm a member of Program B. (Question 8)\\
    \textit{Agent}: Perfect! Your loyalty program membership will be taken into account for additional perks. Based on your preferences, I recommend renting a car from Company D with the extra insurance coverage. We'll make sure you get the best deal as a Program B member. Is there anything else you'd like to add or any other specific requirements? (Recommendation)
     \\ 
     \textbf{Input}: Given a task flow:\\
    ```\\
    \{dialogue flow\}\\
    ```\\
    Please create a natural conversation between a user and an agent using the given task flow. \\
    Include the question number at the end of each utterance. \\
    \textbf{Output}: \{generated dialogue\} 
    \end{tcolorbox}
    \caption{The prompt and in-context example for dialogue generation with dialogue flow.
    \label{tab:dialog_prompt}}
\end{figure*}

\section{\datasets Simulation Framework}
\label{sec:method}

\autoref{fig:overview} provides an overview of our framework that includes three steps: (1) generating task plan based on task instruction, (2) sampling dialogue flows from task plan, and (3) generating diverse and coherent dialogues based on sampled flows.  


\paragraph{Step 1: Task Plan Generation.}
Given a task instruction and one in-context example, the LLM planner is prompted to generate a decision tree-structured task plan that outlines multiple trajectories, as shown in \autoref{tab:plan_prompt}.
Since we aim at prompting the diversity of task logic at the dialogue-level, we adopt the decision tree-structured task plan to cover diverse trajectories.
Each step in the plan has two components: (1) a system \textit{action} which collects user information in order to fulfill user requests; (2) a set of \textit{values} which guides or constrains the system when performing certain action. 

We categorize our system actions by viewing them as different types of nodes in the decision tree, in order to reflect different structures of generated task plans.
Here are the details of each system action:
\begin{itemize}[noitemsep,topsep=2pt,leftmargin=*]
    \item \textit{Yes/No Questions}: This system action collects user information or feedback by asking binary choice questions, which leads to a switch to different branches for task completion.
    \item \textit{Multiple Choice Questions}: This system action collects user information or feedback by asking multiple choice questions, which leads to a continuation to the next valid action.
    \item \textit{User Information Requests}: This system action collects user information or feedback by asking user entering texts, which also leads to a continuation to the next valid action.
    \item \textit{Recommendations}: This system action provides final system recommendation to fulfill the user request, which marks the end of the flow.
\end{itemize}

\paragraph{Step 2: Dialogue Flow Sampling.}
For each task plan, the depth-first-search is applied to extract all valid trajectories, where each trajectory represents a dialogue flow.
At each step, we sample one value under the current action, and proceed to the next step based on the selected value. 
The search continues until reaching the end of the task plan.

Furthermore, to enhance dialogue system to address abnormal user requests, such as inquiries for products or services not offered by the system, or early ending conversations before task completion, we introduce two additional types of error-handling flows to manage these scenarios:
\begin{itemize}[noitemsep,topsep=2pt,leftmargin=*]
    \item \textit{Out-of-scope Request Flow}: This flow is designed to handle requests that violate system constraints. 
    To implement it, we prompt the LLM synthesizer to simulate user providing inputs that violate the system constraints, and asking the system to throw out error message and guide the user to enter valid inputs, as shown in \autoref{tab:dialog_prompt_ood}.
    \item \textit{Early-stop Conversation Flow}: This flow addresses scenarios where the user decides to end the conversation before the task is completed. To implement it, we prompt the LLM synthesizer to simulate user rejecting system recommendations and ending the conversation before task completion, as demonstrated in \autoref{tab:dialog_prompt_end}.
\end{itemize}

\paragraph{Step 3: Dialogue Generation.}
Given a dialogue flow, the LLM synthesizer generates a multi-turn dialogue, where each turn is associated with each step in the dialogue flow, as demonstrated in \autoref{tab:dialog_prompt}.
An automatic filter is then applied to filter out low-quality dialogues with repetitive utterances or utterances that are not associated with any step in the dialogue flow.

\begin{table*}[t]
    \centering
        \resizebox{.78\textwidth}{!}{%
    \begin{tabular}{l|rrrr}
    \toprule
     & \textbf{GPT-3.5-turbo} & \textbf{Mixtral-8x7B-instruct} & \textbf{Falcon-40B-instruct} & \textbf{MPT-30B-chat} \\
    \midrule
    \# Domains & 15 & 15 & 15 & 15  \\
    \# Task Instructions & 130 & 130 & 130 & 130  \\
    \# Plans & 110 & 98 & 83 & 103 \\
    \# Flows & 1,192 & 1,222 & 765 & 781 \\
    ~~- \# Normal Flows & 916 & 1,118 & 559 & 578 \\
    ~~- \# Error Handling Flows & 276 & 104 & 206 & 203 \\
    \# Dialogues & 1,192 & 1,222 & 765 & 781 \\
    \# Utterances & 11,015 & 8,784 & 7,448 & 7,729 \\
    \midrule
    Avg. Plan Steps & 9.33 & 8.68 & 10.95 & 9.07 \\
    Avg. Flow Steps & 7.78 & 6.29 & 9.73 & 8.10 \\
    Avg. Dialogue Turns & 9.24 & 7.18 & 9.73 & 9.89 \\
    \bottomrule
    \end{tabular}}
    \caption{Statistics of \dataset~generated from four LLMs.}
    \label{tab:dataset_statis_llm}
\end{table*}

\section{Experiments}
This section introduces the dataset construction details in \autoref{sec:dataset_construct}, the intrinsic data diversity and quality evaluation in \autoref{sec:intrinsic_eval}, and the extrinsic data effectiveness evaluation for building LLM agents in \autoref{sec:extrinsic_eval}.

\subsection{Dataset Construction}
\label{sec:dataset_construct}
\paragraph{Seed Task Instruction Construction.}
To cover a wide range of real-world tasks, we first build a seed task pool by selecting popular domains, and constructing task instructions under each domain with GPT-3.5-turbo \citep{gpt_3.5}. 
The construction details are as follows.
First, we selecting 15 domains based on \citet{zhang-etal-2024-dialogstudio}, including bank, insurance, travel, car rental, restaurant, shopping, doctor, event, apartment, meeting, ride sharing, payment, weather, calendar, navigate.
Next, we prompt GPT-3.5-turbo to generate task instructions under each domain.
Concretely, we break the task instruction generation into two steps: (1) service name generation, where we prompt GPT-3.5-turbo with ``Please generate 20 common services in \{domain name\} domain for task-oriented dialogue systems''; (2) intent description generation, where we prompt GPT-3.5-turbo with ``Please convert the above services into user intents with intent descriptions''.
Finally, we manually select 130 task instructions across the above 15 domains.

\begin{table}[t]
    \centering
    \small
    \begin{tabular}{l|rrr}
    \toprule
     & \textbf{Train} & \textbf{Test} & \textbf{Total} \\
    \midrule
    \# Domains & 10 & 5 & 15  \\
    \# Task Instructions & 100 & 30 & 130  \\
    \# Plans & 293 & 101 & 394 \\
    \# Flows & 3,229 & 657 & 3,886 \\
    \# Dialogues & 3,229 & 657 & 3,886 \\
    \# Utterances & 29,342 & 5,634 & 34,976 \\
    \midrule
    Avg. Plan Steps & 9.66 & 8.16 & 9.39 \\
    Avg. Flow Steps & 7.98 & 6.78 & 7.76 \\
    Avg. Dialogue Turns & 9.03 & 7.89 & 8.83 \\
    \bottomrule
    \end{tabular}
    \caption{Statistics of our \datasets dataset, where domains in the test set have no overlap with the train set.}
    \label{tab:dataset_statis}
    \vspace{4mm}
\end{table}

\paragraph{Dialogue Simulation.}
We initiate the simulation by inputting task instructions into our proposed framework. During this process, the same LLM initializes both the planner and synthesizer components. To enhance the diversity of the generated task plans, we deploy 4 LLMs to conduct the simulation independently, including GPT-3.5-turbo, Falcon-40B-instruct \citep{falcon40b}, MPT-30B-chat \citep{MosaicML2023Introducing} and Mistral-8x7B-instruct \citep{jiang2024mixtral}.
Then, we combines all generated data from 4 LLMs to construct our \datasets dataset. The detailed data statistics are provided in \autoref{tab:dataset_statis_llm} and \autoref{tab:dataset_statis}. 

In total, we collect 394 task plans across 15 domains from 4 LLMs. As demonstrated in \autoref{tab:dataset_statis}, our \datasets dataset includes 3,886 flows with an average steps of 7.76, and generate 3,886 dialogues based on the flow with an average number of turns of 8.83.
In comparison with the human-annotated ABCD dataset \citep{chen-etal-2021-action} that has fine-grained dialogue flow annotations, our \datasets dataset has longer dialogue flows and also introduces the error handling flows to simulate real-world conversation scenarios.

\subsection{Intrinsic Evaluation}
\label{sec:intrinsic_eval}
After dataset construction, we evaluate both the diversity and quality of the \datasets dataset to ensure that our simulation framework is capable of generating dialogues of high diversity and high quality.

\begin{figure*}[t]
  \centering
  \subfigure[Task domain distribution]{
    \includegraphics[width=0.28\linewidth]{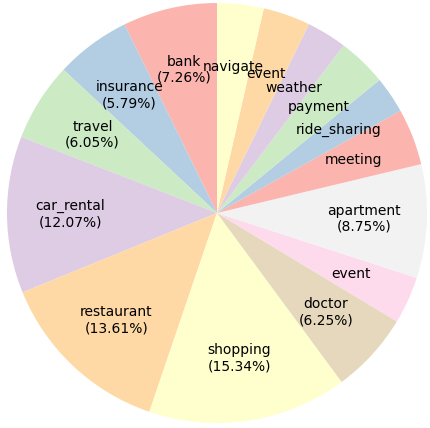}
  }\hfill
  \subfigure[Plan action distribution]{
    \includegraphics[width=0.28\linewidth]{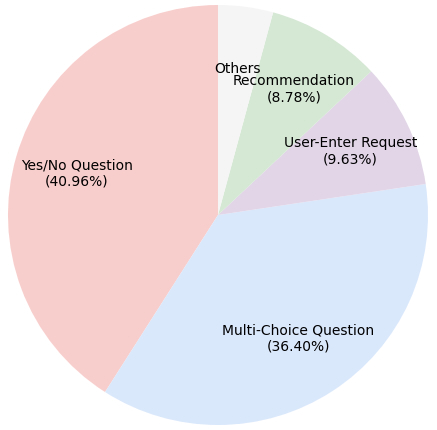}
  }\hfill
  \subfigure[Flow type distribution]{
  \includegraphics[width=0.28\linewidth]{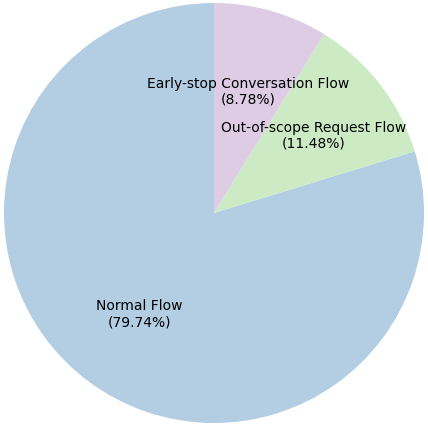}
  }
  \vspace{-3mm}
  \caption{
  Diverse domain, plan action type and flow type distribution of our \datasets dataset.
\label{fig:diversity_plots}}
\end{figure*}

\begin{figure*}[t]
  \centering
  \subfigure[35 dialogue flows from "Agenda creation" plan generated by Mistral-8x7B-instruct.]{
    \includegraphics[width=0.5\linewidth]{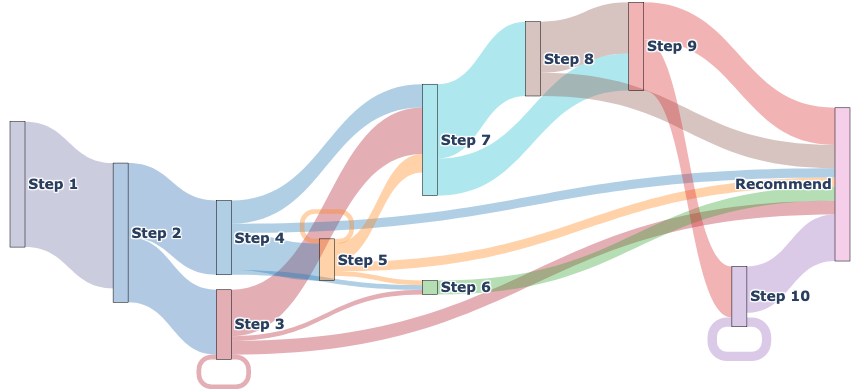}
  }\hfill
  \subfigure[18 dialogue flows from "Long-term car rental" plan generated by GPT-3.5-turbo.]{
    \includegraphics[width=0.45\linewidth]{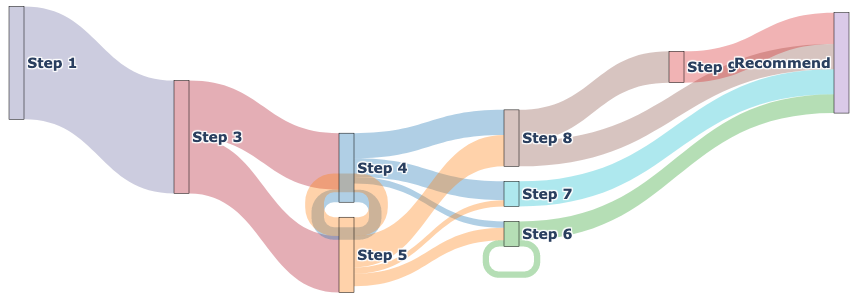}
  }
  \vspace{-3mm}
  \caption{
  Sankey diagram showing diverse dialogue flows of our \datasets dataset. The height of a node shows the number of flows passing through the node.
\label{fig:flows_plots}}
\end{figure*}

\paragraph{Data Diversity.}
We assess the diversity of \datasets from two perspectives: plan diversity and flow diversity.
For plan diversity, we apply multiple LLMs to generate various task plans across 15 task domains, as shown in \autoref{fig:diversity_plots} (a), each reflecting different task logics.
Additionally, the generated task plans cover different types of system actions, as shown in \autoref{fig:diversity_plots} (b) and \autoref{tab:plan_diversity}, each fulfilling different user requests.
For flow diversity, we not only sample diverse dialogue flows from the generated task plan, but also introduce two error-handling flows to address abnormal user requests as shown in \autoref{fig:diversity_plots} (c).
Furthermore, different task plans lead to diverse dialogue flows. \autoref{fig:flows_plots} (a) and \autoref{fig:flows_plots} (b) display diverse dialogue flows sampled from the "Agenda creation" plan and the "Long-term car rental" plan respectively.
This variety in dialogue flows also enriches the diversity of the multi-turn dialogues, as each dialogue flow controls the generation of each multi-turn dialogue.


\begin{table}[t]
    \centering
    \small
    \begin{tabular}{@{}l@{\hskip 2mm}|c@{\hskip 2mm}|c}
    \toprule
    & \dataset & ABCD  \\
    & (flow by auto) & (flow by human) \\
    \midrule
    \textbf{Plan Quality} & 0.9835 & 0.9943 \\
    \textbf{Plan Coherence} & 0.9182 & 1.0000 \\
    \textbf{Flow Coherence} & 0.9220 & 1.0000 \\
    \textbf{Dialogue Faithfulness} & 0.8718 & 0.8011 \\
    \bottomrule
    \end{tabular}
    \caption{Data quality evaluation on \datasets and ABCD. GPT-4 evaluation results on all data pairs <plan, flow, dialogue> from \datasets (full) and ABCD (test).}
    \label{tab:human_eval_full}
    \vspace{4mm}
\end{table}

\begin{table*}[t]
    \centering
    \small
    \begin{tabular}{@{}l@{\hskip 2mm}rrrrr}
    \toprule
     \textsc{Simulation LLMs} & \textbf{\% Y/N Question} & \textbf{\% Multi. Question} & \textbf{\% Info. Request} & \textbf{\% Recommend} & \textbf{\% Others} \\
    \midrule
    GPT-3.5-turbo & \textbf{43.58\%} & 34.91\% & 9.36\% & 9.29\% & 2.83\% \\
    Mistral-8x7B-instruct-v0.1 & 42.49\% & 33.34\% & 6.63\% & \textbf{9.78\%} & \textbf{7.74\%} \\
    Falcon-40B-instruct & 40.84\% & 36.44\% & \textbf{13.91\%} & 6.05\% & 2.73\% \\
    MPT-30B-chat & 34.47\% & \textbf{43.50\%} & 9.89\% & 9.40\% & 2.71\% \\
    \midrule
   Distribution in ICL example & 36.36\% & 45.45\% & 9.09\% & 9.09\% & 0.00\% \\
    \bottomrule
    \end{tabular}
    \caption{System action distributions in \dataset. Given the same task instruction and in-context learning (ICL) example, we show the action distribution in the generated task plans from each LLM.
    Different LLM exhibits distinct bias when generating a certain type of actions. For instance, GPT-3.5-turbo prefers yes/no questions and MPT-30B-chat favors multiple choice questions.}
    \label{tab:plan_diversity}
    \vspace{-2mm}
\end{table*}

\paragraph{Data Quality.}
We evaluate the quality of the \datasets dataset through four perspectives: plan quality, plan coherence, flow coherence and dialogue faithfulness. 

We first manually evaluate 100 sampled simulation data pair <plan, flow, dialogue> from \datasets based on the following questions:
\begin{enumerate}[noitemsep,topsep=2pt,leftmargin=*]
    \item \textbf{Plan quality}: Are all steps in the plan relevant to the task instruction? 
    \item \textbf{Plan coherence}: Is the plan coherent without any repetitive steps?
    \item \textbf{Flow coherence}: Is the flow coherent and without any self-contradictory steps?
    \item \textbf{Dialogue faithfulness}: Is the dialogue faithful to the flow (matching each step in the flow)? 
\end{enumerate}
Five annotators give 0 (bad) or 1 (good) score to each question, and achieve 0.5780 Fleiss's kappa score \citep{fleiss1971measuring}, indicating a moderate inter-annotator agreement. 
For the 100 manually evaluated samples, we obtained an average score of 0.93 for plan quality, 0.866 for plan coherence, 0.878 for flow coherence and 0.836 for dialogue faithfulness. 
Human evaluation results suggest that our proposed method can produce high quality task plans, coherent flows and faithful dialogues.

Then we use an GPT-4 evaluator to scale up the evaluation on all data in \datasets and the test set of ABCD, which is a human-annotated task-oriented dataset with fine-grained dialogue flow annotations. 
The GPT-4 evaluator achieves 0.92 accuracy on our 100 human evaluation data, and the prompt details are provded in \autoref{tab:data_eval_prompt}.
\autoref{tab:human_eval_full} shows that although \datasets is completely constructed by LLMs, it manages to achieve comparable quality with the human-curated dataset in terms of plan quality, flow coherence and dialogue faithfulness.

\begin{table*}[th]
    \centering
    \fontsize{7}{9}\selectfont
    \small
    \begin{tabular}{ll|rrr|rrr}
    \toprule
    & & \multicolumn{3}{c|}{\datasets \textsc{test} }  & \multicolumn{3}{c}{\textsc{ABCD test (human)}} \\
    \textbf{Setup} & \textbf{Model} & \textbf{Action} & \textbf{Value} & \textbf{Accuracy} & \textbf{B-Slot} & \textbf{Value} & \textbf{Accuracy} \\
    \midrule
    \midrule
    \multirow{4}{*}{0-shot} & GPT-4 & 73.72\% & 77.99\% & 73.29\% & 70.46\% & 87.82\% & 70.26\% \\
    & Mistral-large-2402 & 66.67\% & 60.29\% & 55.34\% & 58.68\% & 78.04\% & 57.29\% \\
    & Mistral-7B-instruct-v0.2 & 39.10\% & 29.70\% & 23.50\% & 43.91\% & 65.27\% & 41.92\% \\
    & OpenLlama-7B-instruct & 14.10\% & 14.32\% & 9.19\% & 27.74\% & 51.70\% & 22.55\% \\
    \midrule
    \multirow{4}{*}{3-shot} & GPT-4 & 79.70\% & 82.26\% & 79.27\% & 73.25\% & 86.03\% & 73.05\% \\
    & Mistral-large-2402 & 72.86\% & 76.50\% & 72.44\% & 62.67\% & 79.04\% & 61.48\% \\
    & Mistral-7B-instruct-v0.2 & 39.53\% & 38.46\% & 30.56\% & 43.11\% & 65.67\% & 42.32\% \\
    & OpenLlama-7B-instruct & 22.01\% & 24.36\% & 19.44\% & 32.33\% & 56.69\% & 31.14\% \\
    \midrule
    \midrule
    Fine-tune & Mistral-7B-instruct-v0.2 & \textbf{84.40\%} & \textbf{85.47\%} & \textbf{84.40\%} & 84.83\% & 92.42\% & {84.63\%} \\
     (with \dataset) & OpenLlama-7B-instruct & 73.29\% & 74.79\% & 73.08\% & \textbf{87.62\%} & \textbf{93.41\%} & \textbf{86.63\%} \\
    \midrule
    Fine-tune (w/o flow) & Mistral-7B-instruct-v0.2 & 39.96\% & 41.45\% & 35.90\% & 17.96\% & 50.10\% & 13.17\% \\
    (Ablation) & OpenLlama-7B-instruct & 36.11\% & 37.61\% & 25.43\% & 26.35\% & 53.89\% & 24.95\% \\
    \bottomrule
    \end{tabular}
    \caption{Next action prediction performance on the test set of \dataset~and ABCD, where both fine-tune setups use the training set of \dataset~to fine-tune models. For \dataset~and ABCD, the domains and tasks in test set have no overlap with the training set.}
    \label{tab:ours_ood}
    \vspace{-5mm}
\end{table*}

\subsection{Extrinsic Evaluation}
\label{sec:extrinsic_eval}
Previous studies \citep{zhou2023lima,mekala-etal-2024-smaller} found that a few thousand high quality training data can significantly improve the instruction-tuned language models.
Therefore, we fine-tune instruction-tuned models with 7B parameters in next action prediction tasks, in order to evaluate the effectiveness of the \datasets dataset in building task-oriented LLM agents.

\paragraph{Data Format Setup.}
We convert original datasets to the instruction-tuning data format following \citet{hattami2023workflow}.
For the next action prediction task, the model's input is a dialogue context $C_{t-1}=\{u_1, \dots, u_{t-1}\}$ and a dialogue flow $F=\{a_1[v_1^1], \cdots, a_T[v_T^1]\}$, where $u_i$ is the $i$-th utterance, $a_i$ is the action for the $i$-th utterance, and $[v_i^1, \dots, v_i^m]$ is the value set for action $a_i$. The output of the model is the next system action $a_t[v_t^1, \cdots, v_t^n]$ at the $t$-th utterance.

\paragraph{Evaluation Datasets.}
Since there is lack of fine-grained annotations of dialogue flows in existing dialogue datasets, we choose our \datasets and ABCD as the benchmark test sets to evaluate the dialogue understanding ability of LLMs.  

For \dataset, we obtain 8906 instruction-tuning training data and 468 instruction-tuning test data.
For ABCD, we sample 290 instruction-tuning training data and 500 instruction-tuning test data to mimic the challenging use case where only a few human-annotated flow data is available.

\paragraph{Baseline Models.} 
We evaluate the dialogue understanding ability of 4 models in the next action prediction task, including two strong proprietary LLMs, GPT-4 and Mistral-large, and two open-sourced instruction fine-tuned 7B models, Mistral-7B-instruct-v0.2 \citep{jiang2023mistral} and OpenLlama-7B-instruct \citep{open_llama_7b}.

\paragraph{Evaluation Metrics.}
The evaluation metrics are the \textbf{Action} or \textbf{B-Slot} accuracy on predicting the correct action $a_t$, \textbf{Value} accuracy on predicting the correct values $[v_t^1, \cdots, v_t^n]$, and \textbf{Joint Accuracy} on predicting both the correct action and values $a_t[v_t^1, \cdots, v_t^n]$ at the $t$-th utterance.

\paragraph{Zero-shot Setup.}
The zero-shot setup has no training data provided in the input, but a single formatting example is included to guide the model generate expected output format.
\autoref{tab:zero_shot_prompt_ours} and \autoref{tab:zero_shot_prompt_abcd} demonstrate the prompt in the zero-shot setup on \datasets and ABCD respectively.

\paragraph{Few-shot Setup.}
The few-shot setup has three in-context examples sampled from the training set, which are used to guide the model generate expected output format.
\autoref{tab:few_shot_prompt_ours} and \autoref{tab:few_shot_prompt_abcd} demonstrate the prompt and in-context examples in the few-shot setup on \datasets and ABCD respectively.

\paragraph{Fine-tuning Setup.}
The fine-tuning setup fine-tunes two 7B models using the \datasets dataset and compares their performance with two advanced LLMs under zero-shot and few-shot settings. 
For both \datasets and ABCD, the domains and tasks in the test set have no overlap with the training set.
When testing the fine-tuned LLMs on ABCD test, we add 290 training data from ABCD into our \datasets training set, in order to guide the model learning the ABCD's output format.
We perform LoRA fine-tuning \citep{hu2021lora} to train 7B models.
For each mdoel, we set the max sequence length to 1024, lora rank to 16, lora alpha to 32, learning rate to $2e-4$, batch size per device to 8, and training epochs to 6.
All experiments are conducted on 4 NVIDIA A10G with 24GB GPU memory.

\paragraph{Result Analysis.}
As shown in the fine-tuning section of \autoref{tab:ours_ood}, the two 7B models demonstrated significant improvements in joint accuracy across both test sets. Notably, the fine-tuned Mistral-7B-instruct-v0.2 outperforms the 3-shot GPT-4 by a large margin in both test sets, underscoring the high quality of training data provided by \dataset. This confirms that our dataset effectively enhances model performance in unseen domains and tasks.

The 0-shot section in \autoref{tab:ours_ood} presents the model performance when no training data is provided.
We observe that GPT-4 and Mistral-large achieves similar performance under both our \dataset~and ABCD, which indicates our \dataset~is as challenging as ABCD to current LLMs.
For 7B models, our \dataset~is even more challenging than ABCD.
These results indicate that current LLMs still find it difficult to understand the dialogue flow in task-oriented dialogues. 

The 3-shot section in \autoref{tab:ours_ood} presents the model performance when 3 training data are included as in-context examples.
We find that all models get small improvements in joint accuracy on our \dataset~and ABCD, but the 7B models still find our \dataset~more challenging than ABCD.
Besides, the two LLMs still find it challenging to accurately predict the next action and values, where the 3-shot GPT-4 only achieves 79.27\% joint accuracy, and the 3-shot Mistral-large only achieves 72.44\% joint accuracy in \dataset.
This again highlights that fine-tuning small language models on \datasets can more effectively achieve better performance in the challenging dialogue state understanding task.

\paragraph{Ablation Study.}
To determine the impact of dialogue flows on next action prediction accuracy, we conduct an ablation study by removing all dialogue flows from the \dataset. We then fine-tune two 7B models using the modified dataset with the same training configurations. The results, as shown in the last two lines of \autoref{tab:ours_ood}, indicate a significant drop in performance when dialogue flows are excluded. This suggests that the models heavily rely on dialogue flows to inform their predictions in next action prediction tasks.
\vspace{-1mm}
\section{Conclusions}
\vspace{-1mm}
In this work, we propose a novel data simulation framework to synthesize diverse task-oriented dialogues that follow task logic and constraints for building LLM agents.
We leverage the proposed framework to construct the \datasets dataset with 3.8K fine-grained dialogue flow annotations across 15 domains.
Empirical experiments show that the \datasets dataset achieves comparable data quality with human annotations, and significantly improves 7B models' performance in the next action prediction task, outperforming strong LLMs such as GPT-4.

\section{Limitations}
While our framework successfully generates diverse task-oriented dialogues across 15 domains, the chosen domains may not comprehensively represent the broader range of possible scenarios encountered in real-world applications. 
The diversity and accuracy of the generated dialogues heavily rely on the underlying planning capabilities of the LLMs employed. Although we utilize state-of-the-art models like Mistral-8x7B-instruct and GPT-3.5-turbo, these models are still susceptible to biases present in their training data or inherent limitations in understanding nuanced user intents. 
The inference of some LLMs requires large GPU resources, and future research on memory-efficient inference may enable 100B+ LLMs.
In addition, the current method only generates text-based dialogues, future research may further explore different data sources for task plan generation and dialogue simulation, such as images, graphs, and tabular data.

\section{Ethical Considerations}
We honor the ethical code in the ACL Code of Ethics.
Our simulation datasets respect the copyrights of original LLM authors.
During the data evaluation process, the privacy of all human annotators is respected.
The dataset collection process and conditions are detailed in the paper, and the characteristics of the dataset are described in the paper as well.
Our work has no potential harm to marginalized or vulnerable populations.
Our datasets do not contain any identity characteristics (e.g., gender, race, ethnicity), and will not have ethical implications of categorizing people.

\bibliography{anthology,custom}

\clearpage
\appendix

\onecolumn
\section{Appendix}

\begin{figure*}[th]
    \centering
    \begin{tcolorbox}[colback=gray!10,colframe=black,fontupper=\small,title=Prompt for Task Plan Generation]
\textbf{Instruction}: Please generate a decision tree for a given task that can achieve its goal.\\
     Make sure that the generated content is logically coherent, diverse and non-repetitive. \\
     Below is an example of this task. \\
\textbf{Input}: Please provide an example of decision tree for ``Explore the various car rental services offered''. \\
     \textbf{Output}:  
     1. Are you looking for a specific type of car?\\
       - Yes: Proceed to question 2.\\
       - No: Proceed to question 3.\\
    2. What type of car are you looking for?\\
       - Economy car\\
       - Sedan\\
       - SUV\\
       - Luxury car\\
    3. Do you have a preferred car rental company?\\
       - Yes: Proceed to question 4.\\
       - No: Proceed to question 5.\\
     4. Which car rental company would you prefer?\\
       - Company A\\
       - Company B\\
       - Company C\\
       - Company D\\
    5. What is your budget for car rental?\\
       - Low budget\\
       - Moderate budget\\
       - High budget\\
    6. Do you need any additional services?\\
       - GPS navigation\\
       - Child seat\\
       - Additional insurance coverage\\
       - None\\
    7. Are you a member of any loyalty programs?\\
       - Yes: Proceed to question 8.\\
       - No: Proceed to question 9.\\
    8. Which loyalty program are you a member of?\\
       - Program A\\
       - Program B\\
       - Program C\\
     9. Are there any specific features or requirements you have?\\
       - Yes: Proceed to question 10.\\
       - No: Proceed to recommendation.\\
    10. Please specify your specific features or requirements.\\
    Recommendation: Based on your answers, I would recommend exploring the following car rental services:\\
    - [Car Rental Service 1] \\
    - [Car Rental Service 2] \\
    - [Car Rental Service 3] \\ 
    \textbf{Input}: Please provide an example of decision tree for \{task instruction\}. \\
     \textbf{Output}: \{generated task plan\} \\
\end{tcolorbox}
    \caption{The prompt and in-context example for task plan generation.
    \label{tab:plan_prompt}}
\end{figure*}

\begin{figure*}[h]
    \centering
    \begin{tcolorbox}[colback=gray!10,colframe=black,fontupper=\small,title=Prompt for Dialogue Generation with Out-of-scope Request Flow]
\textbf{Input}: Given a task flow:\\
    ```\\
    3. Do you have a preferred car rental company? - Yes: Proceed to question 4.\\
    4. Which car rental company would you prefer? - Option not available, throw out error message \\
    4. Which car rental company would you prefer? - Company B \\
    6. Do you have any specific requirements for the car?  - No: Proceed to recommendation.\\
    Recommendation \\
    ```\\
    Please create a natural conversation between a user and an agent using the given task flow. \\
    Make sure that the user provides an incorrect answer for questions with invalid options, causing the agent to correct them.  \\
    Include the question number at the end of each utterance.\\
     \textbf{Output}:  \\
     \textit{User}: Hey there! I need to rent a car for an upcoming trip. (Question 3)\\
    \textit{Agent}: Hello! Sure thing. Do you have a preferred car rental company? (Question 3)\\
    \textit{User}: Yeah, I always rent from Company E. (Question 4) \\
    \textit{Agent}: I apologize, but unfortunately, Company E is not an option this time. Could you choose from the available car rental companies? (Question 4)\\
    \textit{User}: Oh, my bad. I guess I'll go with Company B then. (Question 4)\\
    \textit{Agent}: No problem, Company B it is. Lastly, do you have any specific requirements for the car, like additional features or a particular model? (Question 6)\\
    \textit{User}: Nah, nothing specific. Just a good car for my trip. (Question 6)\\
    \textit{Agent}: Got it. Based on your preferences, I recommend a midsize car from Company B. It's a great choice for comfort and reliability. (Recommendation) \\ 
     \textbf{Input}: Given a task flow:\\
    ```\\
    \{out-of-scope request flow\} \\
    ```\\
    Please create a natural conversation between a user and an agent using the given task flow. \\
    Make sure that the user provides an incorrect answer for questions with invalid options, causing the agent to correct them. Include the question number at the end of each utterance.\\
    \textbf{Output}: \{generated dialogue\} \\
    \end{tcolorbox}
    \caption{The prompt and in-context example for dialogue generation with the out-of-scope request flow.
    \label{tab:dialog_prompt_ood}}
\end{figure*}

\begin{figure*}[h]
    \centering
    \begin{tcolorbox}[colback=gray!10,colframe=black,fontupper=\small,title=Prompt for Dialogue Generation with Early-stop Conversation Flow]
\textbf{Input}: Given a task flow:\\
    ```\\
    3. Do you have a preferred car rental company? - Yes: Proceed to question 4. \\
    4. Which car rental company would you prefer? - Ask for recommendation \\
    4. Which car rental company would you prefer? - None of above, end \\
    ```\\
    Please create a natural conversation between a user and an agent using the given task flow.  \\
    Include the question number at the end of each utterance. \\
    Make sure that the agent provides a set of recommendations to the user when asked. \\
    The user chooses none of the above recommendations, and ends the conversation.\\
    \textbf{Output}:  \\
    \textit{Agent}: Good day! Thank you for considering our car rental service. Before we proceed, do you have a preferred car rental company? (Question 3)\\
    \textit{User}: Yes, I do have a preferred car rental company. (Question 3)\\
    \textit{Agent}: Great! Please let us know which car rental company you prefer, and we'll make sure to accommodate your choice. (Question 4) \\
    \textit{User}: Actually, I'm not quite sure which one to choose. Can you give me some recommendations? (Question 4)\\
    \textit{Agent}: Of course! We have a few excellent car rental companies that our customers often prefer. Some popular ones include Avis, Hertz, Enterprise, and Budget. These companies have a wide range of vehicles and offer great service. Would you like more information about any specific one? (Question 4)\\
    \textit{User}: Thank you for the recommendations, but none of those appeal to me. I think I'll consider other options or maybe explore something else. (End of Conversation)\\
    \textit{Agent}: That's completely fine! Feel free to explore other car rental options that suit your preferences better. If you have any further questions or need assistance in the future, don't hesitate to reach out. Have a wonderful day! (End of Conversation) \\ 
    \textbf{Input}: Given a task flow:\\
    ```\\
    \{early-stop conversation flow\} \\
    ```\\
    Please create a natural conversation between a user and an agent using the given task flow.  \\
    Include the question number at the end of each utterance. \\
    Make sure that the agent provides a set of recommendations to the user when asked. \\
    The user chooses none of the above recommendations, and ends the conversation.\\
     \textbf{Output}: \{generated dialogue\}
     \end{tcolorbox}
    \caption{The prompt and in-context example for dialogue generation with the early-stop conversation flow.
    \label{tab:dialog_prompt_end}}
\end{figure*}

\begin{figure*}[h]
    \centering
    \begin{tcolorbox}[colback=gray!10,colframe=black,fontupper=\small,title=Prompt for Data Quality Evaluation]
    \textbf{Plan Quality}: \\
    <prompt> \{task instruction\} </prompt> \\
    <response> \{task plan\} </response> \\
    Consider the text between <prompt> tags as task description, are all questions between <response> tags relevant to the task description? Please answer only with ``yes'' or ``no''.\\

    \textbf{Plan Coherence}: \\
    <prompt> \{task instruction\} </prompt> \\
    <response> \{task plan\} </response> \\
    Consider the text between <prompt> tags as task description, does the text between <response> tags contain repetitive or semantically similar steps? Please answer only with ``yes'' or ``no''.\\

    \textbf{Flow Coherence}: \\
    <prompt> \{task plan\} </prompt> \\
    <response> \{dialogue flow\} </response> \\
    The text between <prompt> tags is the full task plan, and the text between <response> tags is one task flow sampled from the task plan.\\
    Note that `- Option not available, throw out error message` is NOT considered as self-contradictory step!\\
    Does the task flow contain self-contradictory steps? Please answer only with ``yes'' or ``no''. \\

    \textbf{Dialogue Faithfulness}:\\
    <prompt> \{dialogue flow\} </prompt> \\
    <response> \{dialogue\} </response> \\
    The text between <prompt> tags is the task flow, and the text between <response> tags is the dialogue generated based on the task flow.\\
    Compare the task flow and the dialogue to determine whether all the information of the dialogue in present in the task flow or can be inferred from the task flow. \\
    You must answer ”no” if there are any specific details in the dialogue that are not mentioned in the task flow or cannot be inferred from the task flow. Please answer only with ``yes'' or ``no''.\\
    \end{tcolorbox}
    \caption{The prompt for data quality evaluation.
    \label{tab:data_eval_prompt}}
\end{figure*}

\begin{figure*}[h]
    \centering
    \begin{tcolorbox}[colback=gray!10,colframe=black,fontupper=\small,title=Prompt for 0-Shot Next Action Prediction on \datasets]
    \textbf{Prompt}: First, please understand the [context] for this multi-turn conversation; then, please predict the next action for [agent] by selecting the answer from [flow]. Below are a few examples.\\
      \\
    \textbf{Question}: [context] [user] Hi there! I'm interested in opening a new bank account. [agent] Are you looking to open a checking account?  [user] Yes, a checking account would be perfect.  [flow] 1. What type of account are you looking to open? - Checking account; 4. What specific features do you require? - Overdraft facility; 6. What is your budget? - Moderate budget; Recommendation\\
    \textbf{Answer}:  [system] 4. What specific features do you require? - Overdraft facility\\
    
     \textbf{Question}: [context] \{dialogue context\} [flow] \{dialogue flow\} \\
     \textbf{Answer}: [system] \{generated next system action and value\} \\
    \end{tcolorbox}
    \caption{The prompt and in-context example for the next action prediction task in the zero-shot setup on \dataset.
    \label{tab:zero_shot_prompt_ours}}
\end{figure*}

\begin{figure*}[h]
    \centering
    \begin{tcolorbox}[colback=gray!10,colframe=black,fontupper=\small,title=Prompt for 0-Shot Next Action Prediction on ABCD]
    \textbf{Prompt}: First, please understand the [context] for this multi-turn conversation; then, please predict the next action for [agent] by selecting the answer from [flow]. Below are a few examples.\\
      \\
    \textbf{Question}: [context] [customer] would you be able to tell me when my promo code expires? [agent] yes let me look into this [flow] [Search Faq] = search-faq []; [Search Timing] = search-timing []\\
    \textbf{Answer}: [system] search-faq []\\

     \textbf{Question}: [context] \{dialogue context\} [flow] \{dialogue flow\} \\
     \textbf{Answer}: [system] \{generated next system action and value\} \\
    \end{tcolorbox}
    \caption{The prompt and in-context example for the next action prediction task in the zero-shot setup on ABCD.
    \label{tab:zero_shot_prompt_abcd}}
\end{figure*}

\begin{figure*}[h]
    \centering
    \begin{tcolorbox}[colback=gray!10,colframe=black,fontupper=\small,title=Prompt for 3-Shot Next Action Prediction on \datasets]
    \textbf{Prompt}: First, please understand the [context] for this multi-turn conversation; then, please predict the next action for [agent] by selecting the answer from [flow]. Below are a few examples.\\
      \\
    \textbf{Question}: [context] [user] Hi there! I'm interested in getting auto insurance. What types of insurance do you offer?  [agent] Hi there! We offer a variety of auto insurance policies, including liability, collision, and comprehensive coverage. What type of insurance are you looking for?  [user] I'm looking for comprehensive coverage.  [flow] 1. What type of insurance are you looking for? - Auto insurance; 4. What is your budget for insurance? - High budget; 6. Which loyalty program are you a member of? - Program A; 8. Please specify your specific features or requirements.; Recommendation\\
    \textbf{Answer}: [system] 4. What is your budget for insurance? - High budget\\
    \\
    \textbf{Question}: [context] [user] Hi there! I'm planning a trip to a foreign country and I'm considering renting a car. Can you help me with that?  [agent] Hello! Of course, I'd be happy to assist you. Before we proceed, do you have any experience driving in the specific country you'll be visiting?  [user] No, I don't have any experience driving in that country.  [agent] I understand. In that case, would you be open to hiring a professional driver to ensure a smooth and stress-free experience?  [user] Yes, I think hiring a professional driver would be a good option for me.  [flow] 3. Do you have experience driving in the specific country you will be visiting?    - No: Proceed to question 6.; 6. Are you open to hiring a professional driver?    - Yes: Proceed to question 7.; 7. How long do you need the professional driver for?    - Full day; Recommendation\\
    \textbf{Answer}: [system] 7. How long do you need the professional driver for?    - Full day\\
    \\
    \textbf{Question}: [context] [user] Hi there! I'm interested in booking a meeting room for my upcoming event. Do you have any available?  [agent] Hello! Of course, I'd be happy to assist you with that. Before we proceed, do you have any specific requirements for the meeting room?  [user] Yes, I'm expecting less than 10 participants.  [flow] 1. How many participants are you expecting? - Less than 10; 3. What is the date and time of the meeting? - Date: [Date]; 5. What is the location of the meeting? - Location: [Location]; 7. Please specify your specific requirements for the meeting.; 10. Do you have any specific accessibility requirements for the meeting? - Yes: Proceed to question 11.; 11. Please specify your specific accessibility requirements for the meeting.; 13. Please specify your specific equipment requirements for the meeting.; 16. Do you have any specific security requirements for the meeting? - Yes: Proceed to question 17.; 17. Please specify your specific security requirements for the meeting.; 20. Do you have any specific language requirements for the meeting?; Recommendation\\
    \textbf{Answer}: [system] 3. What is the date and time of the meeting? - Date: [Date]\\
    \\
     \textbf{Question}: [context] \{dialogue context\} [flow] \{dialogue flow\} \\
     \textbf{Answer}: [system] \{generated next system action and value\} \\
    \end{tcolorbox}
    \caption{The prompt and in-context examples for the next action prediction task in the few-shot setup on \dataset.
    \label{tab:few_shot_prompt_ours}}
\end{figure*}

\begin{figure*}[h]
    \centering
    \begin{tcolorbox}[colback=gray!10,colframe=black,fontupper=\small,title=Prompt for 3-Shot Next Action Prediction on ABCD]
    \textbf{Prompt}: First, please understand the [context] for this multi-turn conversation; then, please predict the next action for [agent] by selecting the answer from [flow]. Below are a few examples.\\
      \\
    \textbf{Question}: [context] [agent] hi! [agent] how can i help you? [customer] hi! i need to return an item, can you help me with that? [agent] sure, may i have your name please? [customer] crystal minh [agent] thanks, may i ask the reason for the return? [flow] [Pull Up Account] = pull-up-account ['crystal minh']; [Validate Purchase] = validate-purchase ['cminh730', 'cminh730@email.com', '3348917502']; [Enter Details] = enter-details ['(977) 625-2661']; [Notify Team] = notify-team ['manager']\\
    \textbf{Answer}: [system] pull-up-account ['crystal minh']\\
    \\
    \textbf{Question}: [context] [agent] good afternoon, how can i help you? [customer] just wanted to check on the status of a refund [agent] sure, would you give me your full name or account id [customer] alessandro phoenix [customer] <username> [action] pull-up-account [agent] additional to this you would give me the order id and email [agent] please [customer] <order-id> [customer] <email> [customer] no worries [flow] [Pull Up Account] = pull-up-account ['alessandro phoenix']; [Validate Purchase] = validate-purchase ['aphoenix939', 'aphoenix939@email.com', '7916676427']\\
    \textbf{Answer}: [system] validate-purchase ['aphoenix939', 'aphoenix939@email.com', '7916676427']\\
    \\
    \textbf{Question}: [context] [agent] hello, how can i help you [customer] hello. i have a really cool party coming up. and i need some new clothes asap. i am thinking of ordering them to come by overnight shipping [customer] do you know how much that costs? [flow] [Search Faq] = search-faq []; [Search Pricing] = search-pricing []; [Select Faq] = select-faq ['pricing-3']\\
    \textbf{Answer}: [system] search-faq []\\
    \\
     \textbf{Question}: [context] \{dialogue context\} [flow] \{dialogue flow\} \\
     \textbf{Answer}: [system] \{generated next system action and value\} \\
    \end{tcolorbox}
    \caption{The prompt and in-context examples for the next action prediction task in the few-shot setup on ABCD.
    \label{tab:few_shot_prompt_abcd}}
\end{figure*}

\end{document}